%% file: main.tex
\documentclass{article}

\usepackage[accepted]{icml/icml2021}
\input{packages}

\input{macros}

\icmltitlerunning{Learning by Turning: Neural Architecture Aware Optimisation}

\begin{document}

\twocolumn[
\icmltitle{Learning by Turning: Neural Architecture Aware Optimisation}

\icmlsetsymbol{equal}{*}

\begin{icmlauthorlist}
\icmlauthor{Yang Liu}{equal,aba}
\icmlauthor{Jeremy Bernstein}{equal,cal}
\icmlauthor{Markus Meister}{cal}
\icmlauthor{Yisong Yue}{cal}
\end{icmlauthorlist}

\icmlaffiliation{cal}{Caltech}
\icmlaffiliation{aba}{Abacus.AI}

\icmlcorrespondingauthor{YL}{yang@abacus.ai}
\icmlcorrespondingauthor{JB}{bernstein@caltech.edu}

\icmlkeywords{Machine Learning, ICML}

\vskip 0.3in
]

\printAffiliationsAndNotice{\icmlEqualContribution}

\input{section/0-abstract}
\input{section/1-intro}
\newpage
\input{section/2-related}
\newpage
\input{section/3-derive}
\newpage
\input{section/4-algorithm}
\newpage
\input{section/5-expt}
\newpage
\input{section/8-discuss}

\clearpage
\bibliography{refs}
\bibliographystyle{icml/icml2021}

\clearpage
\appendix
\renewcommand{\thesection}{Appendix \Alph{section}}
\input{section/99-app}

\end{document}

%% file: packages.tex
\usepackage{subfigure}

\usepackage{hyperref}

\usepackage[utf8]{inputenc} %
\usepackage[T1]{fontenc}    %
\usepackage{natbib}
\usepackage{url}            %
\usepackage{booktabs,array} %
\usepackage{multirow}
\usepackage{nicefrac}       %
\usepackage[final]{microtype}      %
\usepackage[final]{graphicx}
\usepackage{enumitem}
\usepackage{amsfonts,amssymb,amsthm,amsmath}
\usepackage{bm}
\usepackage{mathabx}
\usepackage{algorithm,algorithmic}

\usepackage{thmtools,thm-restate}
\usepackage{resizegather}
\usepackage{tikz}
  {
    \theoremstyle{definition}
    \newtheorem{definition}{Definition}
      \theoremstyle{plain}
        \newtheorem{hypothesis}{Hypothesis}

  }

\usepackage[font={small,stretch=1.1},labelfont=it,labelsep=period]{caption}

%% file: macros.tex
\usepackage{dsfont}
\usepackage{stmaryrd}

\usepackage{verbatim}

\newcommand{\eps}{\varepsilon}

\renewcommand{\phi}{\varphi}

\newcommand{\R}{\mathbb{R}}
\newcommand{\Sph}{\mathbb{S}}

\newcommand{\abs}[1]{\vert {#1} \vert}
\newcommand{\norm}[1]{\Vert {#1} \Vert}

\newcommand{\Expect}{\operatorname{\mathbb{E}}}
\newcommand{\Var}{\operatorname{Var}}

\newcommand{\Probe}{\mathbb{P}}

%% file: section/0-abstract.tex
\begin{abstract}
    \normalsize Descent methods for deep networks are notoriously capricious: they require careful tuning of step size, momentum and weight decay, and which method will work best on a new benchmark is a priori unclear. To address this problem, this paper conducts a combined study of neural architecture and optimisation, leading to a new optimiser called \textit{Nero}: the \underline{ne}uronal \underline{ro}tator. Nero trains reliably without momentum or weight decay, works in situations where Adam and SGD fail, and requires little to no learning rate tuning. Also, Nero's memory footprint is $\sim$ square root that of Adam or LAMB. Nero combines two ideas: (1) projected gradient descent over the space of \textit{balanced networks}; (2) neuron-specific updates, where the step size sets the \textit{angle} through which each neuron's hyperplane \textit{turns}. The paper concludes by discussing how this geometric connection between architecture and optimisation may impact theories of generalisation in deep learning.
\end{abstract}

%% file: section/1-intro.tex
\section{Introduction}

Deep learning has brought on a new paradigm in computer science, enabling artificial systems to interact with the world at an unprecedented level of complexity. That said, the core technology relies on various heuristic numerical techniques that are sometimes brittle and often require extensive tuning. A major goal of modern research in machine learning is to uncover the principles underlying learning in neural systems, and thus to derive more reliable learning algorithms.

Part of the challenge of this endeavour is that learning in deep networks is an inherently coupled problem. Suppose that training performance is sensitive to a particular detail of the neural architecture---then it is unclear whether that detail affects the expressivity of the architecture, or just the ability of the descent method to train the architecture.

This observation motivates the \textit{combined study} of architecture and optimisation, and this paper explores several questions at that intersection. First of all:
\begin{itemize}%
    \item[$\left<?\right>$] What is the right domain of optimisation for a neural network's weights? Is it $\R^d$, or something more exotic---such as a Cartesian product of hyperspheres?
\end{itemize}
Typically, optimisation is conducted over $\R^d$, while a careful weight initialisation and a tuned weight decay hyperparameter impose a soft constraint on the optimisation domain. Since normalisation schemes such as batch norm \citep{batchnorm} render the network invariant to the scale of the weights, weight decay also plays a somewhat subtle second role in modifying the effective learning rate. Hyperparameters with this kind of subtle coupling add to the compounding cost of hyperparameter search.

Furthermore, descent methods such as Adam \citep{kingma_adam:_2015} and LAMB \citep{lamb} use either synapse-specific or layer-specific gradient normalisation. This motivates a second question:
\begin{itemize}%
    \item[$\left<?\right>$] At what level of granularity should an optimiser work? Should normalisation occur per-synapse or per-layer---or perhaps, per-neuron?
\end{itemize}

This paper contends that in deep learning, hyperparameters proliferate because of hidden couplings between optimiser and architecture. By studying the above questions, and distilling simple rules that govern optimisation and architecture, this paper aims to make deep learning less brittle---and less sensitive to opaque hyperparameters.

\paragraph{Summary of contributions:}
\begin{enumerate}[topsep=0pt,itemsep=0ex,partopsep=1ex,parsep=1ex]
    \item A new optimiser---\textit{Nero}: the \underline{ne}uronal \underline{ro}tator. Nero performs per-neuron projected gradient descent, and uses $\sim$ square root the memory of Adam or LAMB.
    \item Experiments across image classification, image generation, natural language processing and reinforcement learning, in which Nero's \textit{out-of-the-box} configuration tends to outperform \textit{tuned} baseline optimisers.
    \item Discussion of how the connection between optimisation and architecture relates to generalisation theories, such as PAC-Bayes and norm-based complexity.
\end{enumerate}

%% file: section/2-related.tex
\section{Related work}

This section reviews relevant work pertaining to both neural architecture design and optimisation in machine learning, and concludes with a bridge to the neuroscience literature.

\subsection{Neural Architecture Design}

The importance of wiring constraints for the stable function of engineered neural systems is not a new discovery. One important concept is that of \textit{balanced excitation and inhibition}. For instance,
\citet{Rosenblatt58theperceptron} found that balancing the proportion of excitatory and inhibitory synaptic connections made his perceptron more robust to varying input sizes. Another concept relates to the \textit{total magnitude of synapse strengths}. For example, \citet{Rochester1956TestsOA} constrained the sum of magnitudes of synapses impinging on a neuron so as to stabilise the process of learning. Similar ideas were explored by \citet{von1973self} and \citet{Miller1994TheRO}. These works are early predecessors to this paper's definition of \textit{balanced networks} given in Section \ref{sec:arch}.

Given the resurgence of neural networks over the last decade, the machine learning community has taken up the mantle of research on neural architecture design. Special weight scalings---such as \textit{Xavier init} \citep{pmlr-v9-glorot10a} and \textit{Kaiming init} \citep{he2015delving}---have been proposed to stabilise signal transmission through deep networks. These scalings are only imposed at initialisation and are free to wander during training---an issue which may be addressed by tuning a weight decay hyperparameter. More recent approaches---such as batch norm \citep{batchnorm}---explicitly control activition statistics throughout training by adding extra normalisation layers to the network.

Other recent normalisation techniques lie closer to the work of \citet{Rosenblatt58theperceptron} and \citet{Rochester1956TestsOA}. Techniques that involve constraining a neuron's weights to the unit hypersphere include: weight norm \citep{salimans2016weight}, decoupled networks \citep{Liu2017NIPS,Liu2018DCNets} and orthogonal parameterised training \citep{Liu2021OPT}. Techniques that also balance excitation and inhibition include centred weight norm \citep{huang2017centered} and weight standardisation \citep{Qiao2019MicroBatchTW}.

\subsection{Descent Methods in Deep Learning}

Much classic work in optimisation theory focuses on deriving convergence results for descent methods under assumptions such as \textit{convexity} \citep{boyd} and \textit{Lipschitz continuity of the gradient} \citep{Nesterov2004IntroductoryLO}. These simplifying assumptions are often used in the machine learning literature. For instance, \citet{bottou} provide convergence guarantees for stochastic gradient descent (SGD) under each of these assumptions. However, these assumptions do not hold in deep learning \citep{Sun2019OptimizationFD}.

On a related note, SGD is not the algorithm of choice in many deep learning applications, and heuristic methods such as RMSprop \citep{Tieleman2012} and Adam \citep{kingma_adam:_2015} often work better. For instance, Adam often works much better than SGD for training generative adversarial networks \citep{fromage}. Yet the theory behind Adam is poorly understood \citep{reddi}.

A more recent line of work has explored optimisation methods that make \textit{relative updates} to neural network parameters. Optimisers like LARS \citep{lars}, LAMB \citep{lamb} and Fromage \citep{fromage} make per-layer relative updates, while Madam \citep{madam} makes per-synapse relative updates. \citet{lars} found that these methods stabilise large batch training, while \citet{fromage} found that they require little to no learning rate tuning across tasks.

Though these recent methods partially account for the neural architecture---by paying attention to its layered operator structure---they do not rigorously address the optimisation domain. As such, LARS and LAMB require a tunable weight decay hyperparameter, while Fromage and Madam restrict the optimisation to a bounded set of tunable size (i.e.\ weight clipping). Without this additional tuning, these methods can be unstable---see for instance \citep[Figure 2]{fromage} and \citep[Figure 3]{madam}.

The discussion in the previous paragraph typifies the machine learning state of the art: optimisation techniques that work well, albeit only after hyperparameter tuning. For instance, LAMB is arguably the state-of-the-art relative optimiser, but it contains in total \textit{five} tunable hyperparameters. Since---at least na\"{i}vely---the cost of hyperparameter search is exponential in the number of hyperparameters, the prospect of fully tuning LAMB is computationally daunting.

\subsection{Homeostatic Control in Neuroscience}

Since the brain is a system that must learn stably without hyperparameter do-overs, it is worth looking to neuroscience for inspiration on designing better learning algorithms.

A major swathe of neuroscience research studies mechanisms by which the brain performs homeostatic control. For instance, neuroscientists report a form of homeostasis termed \textit{synaptic scaling}, where a neuron modulates the strengths of all its synapses to stabilise its firing rate \citep{Turrigiano2008TheSN}. More generally, \textit{heterosynaptic plasticity} refers to homeostatic mechanisms that modulate the strength of unstimulated synapses \citep{Chistiakova2009HeterosynapticPI}. \citet{Shen2020ACB} review connections to normalisation methods used in machine learning.

These observations inspired this paper to consider implementing homeostatic control via projected gradient descent---leading to the Nero optimiser.

%% file: section/3-derive.tex
\section{Background Theory}
\label{sec:theory}

In general, an $L$-layer neural network $f(\cdot)$ is a composition of $L$ simpler functions $f_1(\cdot),...,f_L(\cdot)$:
\begin{equation}
    f(x) = f_L\circ f_{L-1} \circ ...\circ f_1(x). \tag{forward pass}
\end{equation}
Due to this compositionality, any slight ill-conditioning in the simple functions $f_i(\cdot)$ has the potential to \textit{compound} over layers, making the overall network $f(\cdot)$ very ill-conditioned. Architecture design should aim to prevent this from happening, as will be covered in Section \ref{sec:arch}

The Jacobian $\partial f/\partial f_l$, which plays a key role in evaluating gradients, also takes the form of a deep product:
\begin{equation}
    \frac{\partial f}{\partial f_l} = \frac{\partial f_L}{\partial f_{L-1}} \cdot \frac{\partial f_{L-1}}{\partial f_{L-2}} \cdot ...\cdot \frac{\partial f_{l+1}}{\partial f_{l}}. \tag{backward pass}
\end{equation}
Therefore, it is also important from the perspective of gradient-based optimisation that compositionality is adequately addressed, as will be covered in Section \ref{sec:opt}.

\subsection{Balanced Network Architectures}
\label{sec:arch}

A common strategy to mitigate the issue of compounding ill-conditioning is to explicitly re-normalise the activations at every network layer. Batch norm \citep{batchnorm} exemplifies this strategy, and was found to improve the trainability of deep residual networks. Batch norm works by standardising the activations across a batch of inputs at each network layer---that is, it shifts and scales the activations to have mean zero and variance one across a batch.

Although batch norm works well, it adds computational overhead to both the forward and backward pass. To explore how far one can get without explicit re-normalisation, the following definitions are useful:

\begin{definition} A neuron is \textit{balanced} if its weight vector $w\in\R^d$ satisfies the following constraints:
    \begin{align}
        \textstyle\sum_{i=1}^d w_i &= 0; \tag{balanced excitation \& inhibition}\\
        \textstyle\sum_{i=1}^d w_i^2 &= 1. \tag{$\ell_2$ constant sum rule}
    \end{align}
\end{definition}
\begin{definition}\label{def:bal-net} A network is \textit{balanced} if all its constituent neurons are balanced.
\end{definition}
As noted by \citet{huang2017centered}, balanced neurons attain some of the properties of batch norm for free. To see this, consider a linear neuron $y = \sum_i w_i x_i$ with inputs $x_i$ that are uncorrelated with mean $\mu$ and variance $1$. Then the output $y$ is standardised:
\begin{align*}
    \Expect [y] &\textstyle= \sum_i w_i \Expect[x_i] = \mu\sum_i w_i = 0; \\
    \Var [y] &\textstyle= \sum_i w_i^2 \Var[x_i] = \sum_i w_i^2 =1.
\end{align*}
While the assumptions on the inputs $x_i$ are unlikely to hold exactly, under more general conditions the constraints may at least \textit{encourage} the standardisation of activation statistics through the layers of the network \citep{brock2021characterizing}.

\subsection{Stable Descent Steps}
\label{sec:opt}

Since a network is trained via perturbations to its parameters, it is important to know what size perturbations are appropriate. Consider an $L$-layer network with weight matrices $W = (W_1,W_2,...,W_L)$ and loss function $\mathcal{L}(W)$. For a perturbation $\Delta W = (\Delta W_1,\Delta W_2,...,\Delta W_L)$, the following definition establishes a notion of stable step size:
\begin{definition}\label{def:stable} Let $\theta_l$ denote the angle between $\Delta W_l$ and $-\nabla_{W_l}\mathcal{L}(W)$. A descent step is \textit{stable} if for all $l=1,...,L$:
\begin{equation}\label{eq:stable}
    \scalebox{0.85}{
    $\displaystyle\frac{\norm{\nabla_{W_l}\mathcal{L}(W+\Delta W)-\nabla_{W_l}\mathcal{L}(W)}_F}{\norm{\nabla_{W_l}\mathcal{L}(W)}_F}<\cos\theta_l.$}
\end{equation}
\end{definition}
\vspace{-.25em}
Or in words: for each layer, the relative change in gradient induced by the perturbation should not exceed the cosine of the angle between the perturbation and the negative gradient.

This definition is useful because a stable descent step is guaranteed to decrease a continuously differentiable loss function $\mathcal{L}(W)$ \citep{fromage}. Still, extracting a stable step $\Delta W$ directly from Inequality \ref{eq:stable} would require first computing extra gradients $\nabla_{W_l}\mathcal{L}(W+\Delta W)$. \citet{fromage} proposed the following model to avoid this:

\begin{definition}\label{def:drt}The loss function obeys \textit{deep relative trust} if for all perturbations $\Delta W = (\Delta W_1,\Delta W_2,...,\Delta W_L)$:
\begin{equation*}
    \scalebox{0.85}{
    $\displaystyle\frac{\norm{\nabla_{W_l}\mathcal{L}(W+\Delta W)-\nabla_{W_l}\mathcal{L}(W)}_F}{\norm{\nabla_{W_l}\mathcal{L}(W)}_F} \leq \prod_{k=1}^L \left(1+\frac{\norm{\Delta W_k}_F}{\norm{W_k}_F}\right) - 1.$}
\end{equation*}
\end{definition}
\vspace{-.25em}
While deep relative trust is based on a perturbation analysis of $L$-layer perceptrons \citep[Theorem 1]{fromage}, the key idea is that its product structure explicitly models the product structure of the network's backward pass. 

The deep relative trust model suggests that a stable descent step should involve small relative perturbations \textit{per layer}. This motivates the layer-wise family of descent methods \citep{lars,lamb}. Still, it is unclear whether layers are the right base object to consider. Perhaps a more refined analysis would replace the layers appearing in Definition \ref{def:drt} with individual \textit{neurons} or even \textit{synapses}.

Small relative perturbations per-synapse were explored by \citet{madam} and found to slightly degrade training performance compared to Adam and SGD. But this paper will explore the per-neuron middle ground:

\begin{definition}\label{def:pnr}
A step of size $\eta>0$ is said to be \textit{per-neuron relative} if for any neuron with weights $w\in\R^d$ and bias $b\in\R$, the perturbations $\Delta w\in\R^d$ and $\Delta b\in\R$ satisfy:
\begin{equation*}
    \norm{\Delta w}_2/\norm{w}_2 \leq \eta \qquad \text{and} \qquad \abs{\Delta b}/\abs{b} \leq \eta.
\end{equation*}
\end{definition}
\vspace{-.25em}

A per-neuron relative update is automatically per-layer relative. To see this, consider a weight matrix $W$ whose $N$ rows correspond to $N$ neurons $w^{(1)},...,w^{(N)}$. Then:
\begin{gather}\label{eq:pnpl}
    \resizebox{0.9\hsize}{!}{
    $\frac{\norm{\Delta W}_F}{\norm{W}_F} = \sqrt{\frac{\sum_{i=1}^N\norm{\Delta w^{(i)}}_2^2}{\sum_{i=1}^N\norm{ w^{(i)}}_2^2}} \leq \sqrt{\frac{\sum_{i=1}^N\eta^2\norm{w^{(i)}}_2^2}{\sum_{i=1}^N\norm{ w^{(i)}}_2^2}} = \eta.$
    }
\end{gather}

%% file: section/4-algorithm.tex
\section{\texorpdfstring{Nero: the \underline{Ne}uronal \underline{Ro}tator}{Nero: the Neuronal Rotator}}

Following the discussion in Section \ref{sec:theory}, this paper will consider an optimisation algorithm that makes \textit{per-neuron relative updates} (Definition \ref{def:pnr}) constrained to the space of \textit{balanced networks} (Definition \ref{def:bal-net}). 

Since a balanced neuron is constrained to the unit hypersphere, a per-neuron relative update with step size $\eta$ corresponds to a pure rotation of the neuron's weight vector by angle $\approx \eta$. To see this, take $\eta$ small in the following picture:\vspace{-.5em}
\input{figures/diagram}
\vspace{-1em}
Hence, this paper proposes \textit{Nero}: the \underline{ne}uronal \underline{ro}tator. Nero's goal is to reduce the burden of hyperparameter tuning by baking architectural information into the optimiser. More concretely, the anticipated advantages are as follows:
\begin{enumerate}[topsep=0pt,itemsep=0ex,partopsep=1ex,parsep=1ex]
    \item Since per-neuron relative updates are automatically per-layer relative by Equation \ref{eq:pnpl}, they should inherit the properties of per-layer updates---in particular, stability across batch sizes \citep{lars} while needing little to no learning rate tuning \citep{fromage}.
    \item Since balanced networks place hard constraints on the norm of a neuron's weights, the need for initialisation tuning and weight decay on these weights is removed.
    \item Gradients are often normalised by running averages, in order to retain relative scale information between successive minibatch gradients \citep{Tieleman2012}. Along with momentum, this is the main memory overhead of Adam and LAMB compared to vanilla SGD. Per-neuron running averages consume $\sim$ square root the memory of per-synapse running averages.
    \item Since normalisation is local to a neuron, no communication is needed between neurons in a layer (unlike for per-layer updates). This makes the optimiser more distributable---for example, a single layer can be split across multiple compute devices without fuss. For the same reason, the Nero update seems more biologically plausible than per-layer optimisers such as LAMB.
\end{enumerate}

There is a significant difference between the implementation of balanced networks in Nero versus prior work. In centred weight norm \citep{huang2017centered} and weight standardisation \citep{Qiao2019MicroBatchTW}, a neuron's underlying weight representation is an \textit{unnormalised} vector $\widetilde{w}\in\R^d$---which is normalised by including the following reparameterisation in the neural architecture:
\begin{equation}\label{eq:cwn}
    \mathtt{normalise}(\widetilde{w}):=\frac{\widetilde{w}-\bm{1}^T\widetilde{w}\cdot\bm{1}/d}{\norm{\widetilde{w}-\bm{1}^T\widetilde{w}\cdot\bm{1}/d}_2},
\end{equation}
where $\bm{1}$ denotes the vector of 1s.

\input{figures/nero}

Since the target of automatic differentiation is still the unnormalised vector $\widetilde{w}$, overhead is incurred in both the forward and backward pass. Moreover, there is a subtle coupling between the step size in additive optimisers like Adam and the scale of the unnormalised weights $\widetilde{w}$---see Section \ref{sec:coupling}.

In contrast, Nero opts to implement balanced networks via projected gradient descent. This is lighter-weight than Equation \ref{eq:cwn}, since duplicate copies of the weights are not needed and the network's backward pass does not involve extra operations. Furthermore, Nero can be used as a drop-in replacement for optimisers like Adam, SGD or LAMB, without the user needing to manually modify the network architecture via the reparameterisation in Equation \ref{eq:cwn}. Note that projected gradient descent arises frequently in machine learning \citep{JMLR:v20:16-607, bai2018proxquant}.

Pseudocode for Nero is provided in Algorithm \ref{alg:nero}. Since Nero normalises gradients via running averages, a Nero update is only approximately per-neuron relative. For brevity, the Adam-style bias correction of the running averages is omitted from the pseudocode. But in the Pytorch implementation used in this paper's experiments, the running averages $\bar{g}_w$ and $\bar{g}_b$ are divided by a factor of $\sqrt{1-\beta^t}$ before the $t$th update. This corrects for the warmup bias stemming from $\bar{g}_w$ and $\bar{g}_b$ being initialised to zero \citep{kingma_adam:_2015}.

While the pseudocode in Algorithm \ref{alg:nero} is presented for \textit{neurons} and \textit{biases}, in the Pytorch implementation the bias update is applied to any parameters lacking a notion of fan-in---including batch norm gains and biases. Typical initialisation scales are $\sigma_b=1$ for gains and $\sigma_b=0.01$ for biases. The Pytorch implementation of Nero defaults to $\sigma_b=0.01$ for any bias parameter initialised to zero.

%% file: figures/diagram.tex
\begin{center}
\hspace{4em}
\begin{tikzpicture}
    \draw[thick,-{latex[scale=2.0]}] (2,2) -- (5,2.5);
    \draw[thick,-{latex[scale=2.0]}] (2,2) -- (5,1.5);
    \draw[thick,-{latex[scale=2.0]}] (5,1.5) -- (5,2.5);
    \draw[thick] (3.25,1.8) arc (-55:55:.25);
    \node at (3.5,1.25) {$\norm{w}_2=1$};
    \node at (3.5,2.75) {$\norm{w+\Delta w}_2=1$};
    \node at (6,2) {$\norm{\Delta w}_2=\eta$};
    \node at (3.6,2) {$\theta$};
\end{tikzpicture}
\end{center}

%% file: figures/nero.tex
\begin{algorithm}[H]
\caption{Nero optimiser. ``Out-of-the-box'' hyperparameter defaults are $\eta=0.01$ and $\beta=0.999$. The constant $\sigma_b\in\R_+$ refers to the initialisation scale of the biases.}
   \label{alg:nero}
\begin{algorithmic}
   \STATE {\bfseries Input:} step size $\eta\in(0,1]$, averaging constant $\beta\in[0,1)$\hspace{-1em}
   \REPEAT
   \FOR{{\bfseries each neuron}}
   \STATE \COMMENT{get weight \& bias gradients $g_w\in\R^n$ \& $g_b\in\R$}\hspace{-1em}
   \STATE \COMMENT{update running averages}
   \STATE $\bar{g}_w^2 \gets \beta\cdot\bar{g}_w^2 + (1-\beta)\cdot\norm{g_w}_2^2$
    \STATE $\bar{g}_b^2 \gets \beta\cdot\bar{g}_b^2 + (1-\beta)\cdot g_b^2$
    \vspace{0.5em}
   \STATE \COMMENT{update weights $w\in\R^n$ and bias $b\in\R$}
    \STATE $w \gets w - \eta\cdot\norm{w}_2/\bar{g}_w \cdot g_w$
    \STATE $b \gets b - \eta \cdot\sigma_b/\bar{g}_b\cdot g_b$
   \vspace{0.5em}
    \STATE \COMMENT{project weights back to constraint set}
    \STATE $w \gets w - \frac{1}{n}\sum_{i=1}^n w_i$
   \STATE $w \gets w / \norm{w}_2$
  \ENDFOR
   \UNTIL{\textbf{converged}}
\end{algorithmic}
\end{algorithm}

%% file: section/5-expt.tex
\section{Experiments}

This section presents experiments intended to demonstrate Nero's key properties. In all figures, the mean and range are plotted over three repeats. For Nero, \textit{out-of-the-box} refers to setting $\eta=0.01$ and $\beta=0.999$. The code for these experiments is available at \href{https://github.com/jxbz/nero}{\texttt{github.com/jxbz/nero}}, and more experimental details are given in \ref{app:expt}.

\subsection{Constraints Help Nero}

To verify that projecting to the space of balanced networks improves the performance of Nero, an ablation experiment was conducted. As can be seen in Figure \ref{fig:ablation}, when training a VGG-11 image classifier on the CIFAR-10 dataset, Nero performed best with both constraints switched on. 

\subsection{Per-Neuron Updates are a Good Middle Ground}

Since \citet{madam} found that per-synapse relative updates led to slightly degraded performance, while per-layer relative updates typically perform well \citep{lars,lamb,fromage}, this section compares per-synapse, per-neuron and per-layer relative updates. In particular, Nero is compared to Madam (per-synapse relative) and LAMB (per-layer relative). 

A VGG-11 model was trained on the CIFAR-10 dataset. Without constraints, the three optimisers performed similarly, achieving $\sim 12\%$ top-1 validation error (Figure \ref{fig:nml}, top). Constraining to the space of balanced networks (Definition \ref{def:bal-net}) improved both Nero and LAMB, but did not have a significant effect on Madam (Figure \ref{fig:nml}, bottom). In both configurations, Nero outperformed Madam and LAMB, demonstrating the viability of per-neuron relative updates.

\begin{figure}
    \input{figures/ablation}
    \input{figures/nml}
    \input{figures/mnist}
\end{figure}

\subsection{The Pitfalls of Reparameterisation}\label{sec:coupling}

Existing implementations of balanced networks (Definition \ref{def:bal-net}) work via the re-parameterisation given in Equation \ref{eq:cwn} \citep{huang2017centered,Qiao2019MicroBatchTW}. This leads to an undesired coupling between the learning rate in optimisers like Adam and the scale of the unnormalised $\widetilde{w}$ parameters.

To verify this, a network with weights normalised by Equation \ref{eq:cwn} was trained to classify the MNIST dataset. The initial weights $\widetilde{w}$ were drawn from $\mathcal{N}(0,\sigma^2)$, and the experiment was repeated for $\sigma=1$ and $\sigma=100$. The Adam optimiser was used for training with a fixed learning rate of $0.01$. As can be seen in Figure \ref{fig:mnist} (left), the training performance was sensitive to the weight scale $\sigma$, despite the fact that a weight normalisation scheme was being used.

The unnecessary scale freedom of reparameterisation can lead to other undesired consequences such as numerical overflow.
Nero completely eliminates this issue by implementing balanced networks via projected gradient descent.

\subsection{Nero Trains Deeper Networks}
Very deep networks are typically difficult to train without architectural modifications such as residual connections \citep{he2016deep} or batch norm \citep{batchnorm}. To test whether Nero enables training very deep models without such modifications, Figure \ref{fig:mnist} (right) shows the results of training a very deep multilayer perceptron (MLP) on the MNIST dataset. Unlike SGD, Adam and LAMB, Nero could reliably train a 100-layer MLP.

\subsection{Nero Works Well Out-of-the-Box}
\label{sec:baselines}

This section probes the versatility and robustness of Nero by comparing its optimisation and generalisation performance with three popular alternatives---SGD, Adam, and LAMB---across six learning problems. The tasks span the domains of computer vision, natural language processing, and reinforcement learning. A wide spectrum of neural architectures were tested---from convolutional networks to transformers.

To make a fair comparison between optimisers, a fair hyperparameter tuning strategy is needed. In this section:
\begin{enumerate}[topsep=0pt,itemsep=0ex,partopsep=1ex,parsep=1ex]
    \item Learning rates were tuned over $\{10^{-4}, 10^{-3}, ..., 10^{0}\}$.
    \item For Adam, LAMB and SGD, the momentum hyperparameter was tuned to achieve good performance on the most complicated benchmark---cGAN training---and then fixed across the rest of the benchmarks. In each case, the best momentum value for cGAN was 0.
    \item $\beta$ in Nero and $\beta_2$ in Adam and LAMB were fixed to $0.999$ across all experiments, as recommended by \citet{kingma_adam:_2015} and \citet{lamb}.
    \item Weight decay was not used in any of the experiments.
\end{enumerate}

The results are collated in Table \ref{table:results}. Nero achieved the best validation performance in every experiment---while the runner-up varied across tasks. What's more, the same learning rate of $\eta=0.01$ was optimal for Nero in five out of six experiments. This means that Nero has strong \textit{out-of-the-box} performance, since Nero's only other hyperparameter was fixed to $\beta=0.999$ across all experiments.

The remainder of this section discusses each experiment in turn. Implementation details are given in \ref{app:expt}.

\input{figures/table}

\begin{figure}
    \input{figures/cGAN}
    \input{figures/cifar}
    \input{figures/wikitext}
\end{figure}

\paragraph{Image synthesis with cGAN} 

Generative Adversarial Network \citep[GAN]{goodfellow2014generative} training is perhaps the most challenging optimisation problem tackled in this paper. Good performance has traditionally relied on extensive tuning: different learning rates are often used in the generator and discriminator \citep{ttur} and training is highly sensitive to momentum \citep[p.~35]{brock2018large}. The class-conditional GAN model in this paper is based on the BigGAN architecture \citep{brock2018large}. This is a heterogeneous network involving a variety of building blocks: convolutions, embeddings, fully connected layers, attention layers, conditional batch norm and spectral norm \citep{miyato2018spectral}. The results are presented in Figure \ref{fig:cgan}.

\vspace{-1em}

\paragraph{Image classification} 

Experiments were run across all baselines on the CIFAR-10 dataset. The networks used were the vanilla, convolutional VGG-11 network \citep{simonyan2014very} and the batch-normalised, residual ResNet-18 network \citep{he2015delving}. The results are presented in Figure \ref{fig:cifar}. ImageNet results using ResNet-50 are presented in Section \ref{sec:imagenet}. Due to limited computational resources, the LAMB and Adam baselines were omitted.

\vspace{-1em}

\paragraph{Natural language processing}

Much recent progress in natural language processing is based on the transformer architecture \citep{transformer}. Transformers process information via layered, all-to-all comparisons---without recourse to recurrence or convolution. This paper experimented with a smaller transformer (19 tensors) trained on the Wikitext-2 dataset, and a larger transformer (121 tensors) trained on WMT2016 English--German translation. The results are presented in Figures \ref{fig:wikitext} and \ref{fig:translation}.

\vspace{-1em}

\paragraph{Reinforcement learning} 

Many reinforcement learning algorithms use neural networks to perform function approximation. Proximal Policy Optimization \citep[PPO]{schulman2017proximal} is one example, and PPO has gained increasing popularity for its simplicity, scalability, and robust performance. This paper experimented with PPO on the Atari Pong video game. The results are presented in Figure \ref{fig:ppo}.

While LAMB failed to train on this task, further investigation revealed that setting LAMB's momentum hyperparameter to 0.9 enabled LAMB to learn. This demonstrates that LAMB is sensitive to the momentum hyperparameter.

\subsection{Nero Can Be Regularised}
\label{sec:imagenet}

This section compares using Nero versus SGD to train a ResNet-50 classifier on the ImageNet dataset. The results are shown in Figure \ref{fig:imagenet}. While out-of-the-box Nero attained the best training error and better validation error than SGD, it performed worse than SGD with tuned weight decay on the validation set. But after fine-tuning the learning rate and adding regularisation, Nero roughly matched SGD with weight decay. In particular, the tuned version of Nero used a learning rate of 0.02 (tuned), a bias scale parameter $\sigma_b=1.0$ (not tuned) and the batch norm gains were regularised towards one using a quadratic penalty.

\begin{figure}[H]
    \input{figures/translation}
    \input{figures/ppo}

    \input{figures/imagenet}
\end{figure}

%% file: figures/ablation.tex
\begin{figure}[H]
    \centering
    \includegraphics[height=0.45\linewidth]{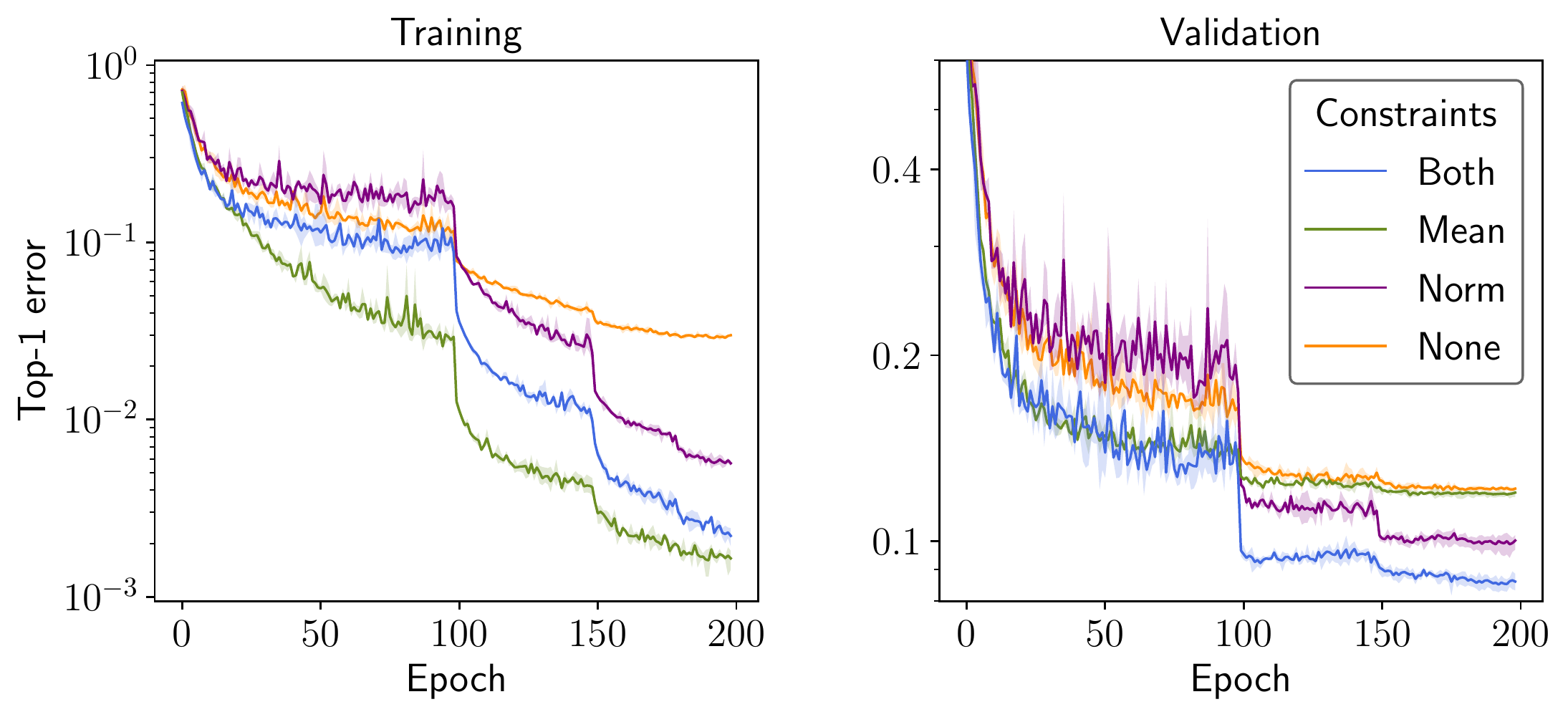}
    \vspace{-.75em}
    \caption{Ablating the \textit{balanced network} constraints. A VGG-11 network was trained on CIFAR-10. The legend denotes which of Nero's constraints were active. \textit{Mean} refers to balanced excitation \& inhibition, while \textit{norm} refers to the $\ell_2$ constant sum rule.}
    \label{fig:ablation}
\end{figure}

%% file: figures/nml.tex
\begin{figure}[H]
    \centering
    \includegraphics[height=0.45\linewidth]{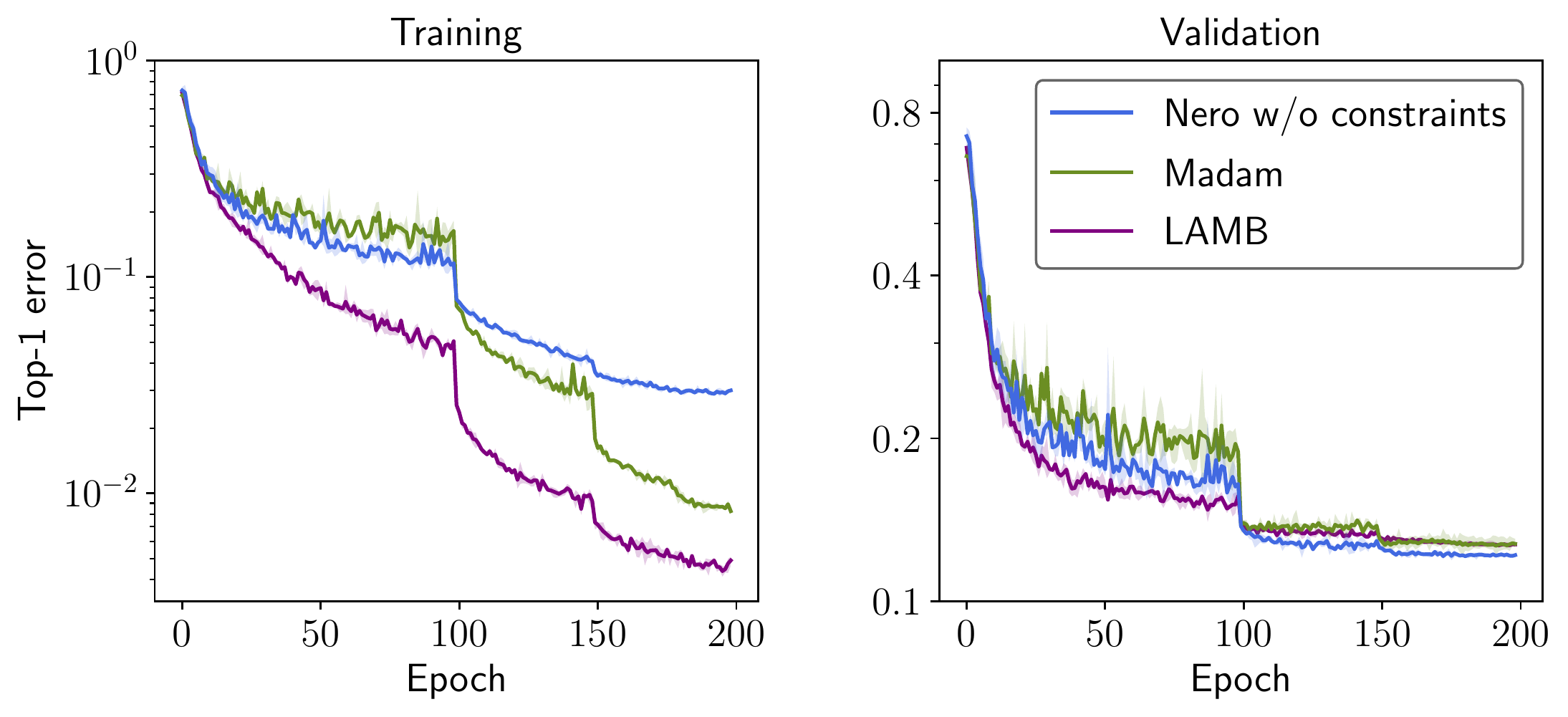} 
    \includegraphics[height=0.45\linewidth]{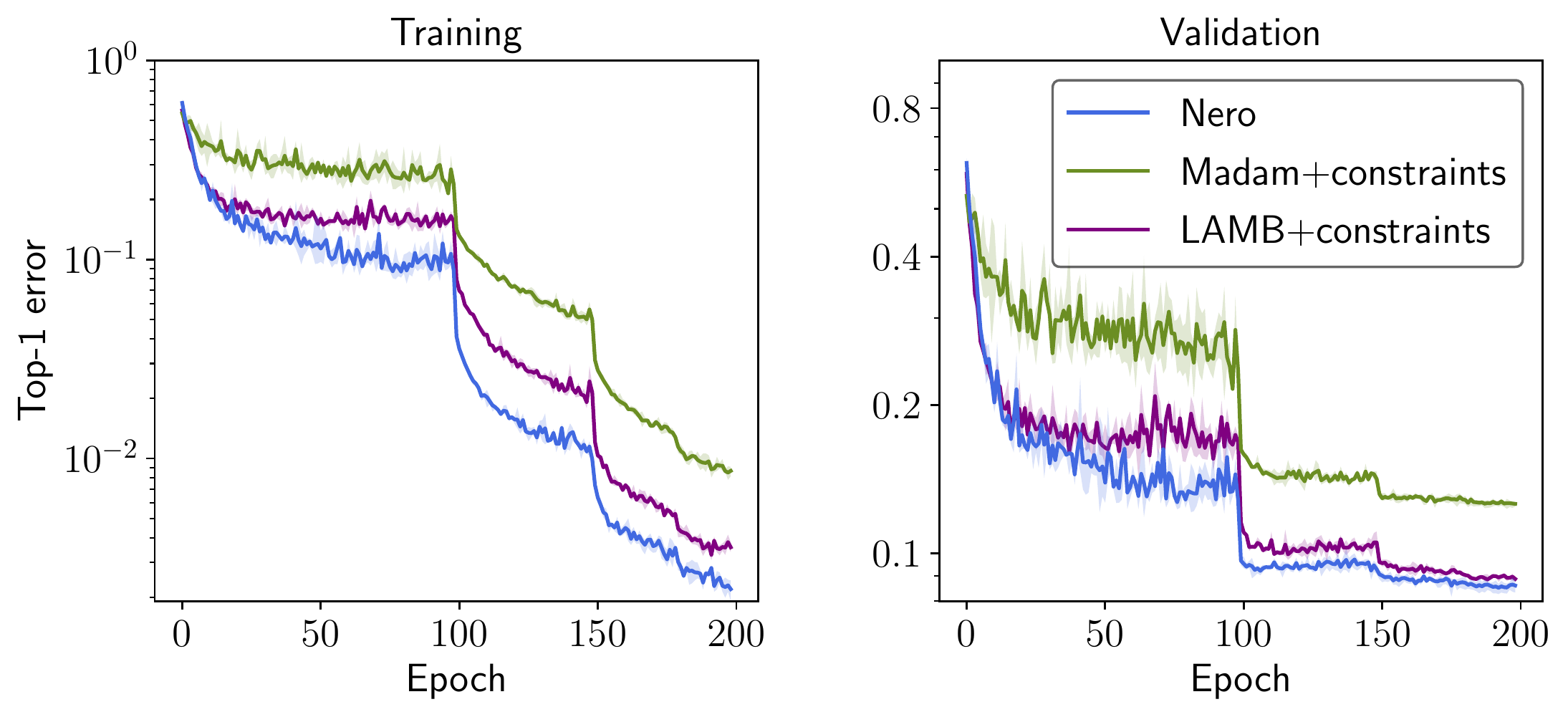} 
    \vspace{-.75em}
    \caption{Comparing per-synapse (Madam), per-neuron (Nero) and per-layer (LAMB) relative updates. A VGG-11 network was trained to classify CIFAR-10. Top: all optimisers \textit{without} balanced network constraints. Bottom: all optimisers \textit{with} constraints.}
    \label{fig:nml}
\end{figure}

%% file: figures/mnist.tex
\begin{figure}[H]
    \centering
    \includegraphics[height=.45\linewidth]{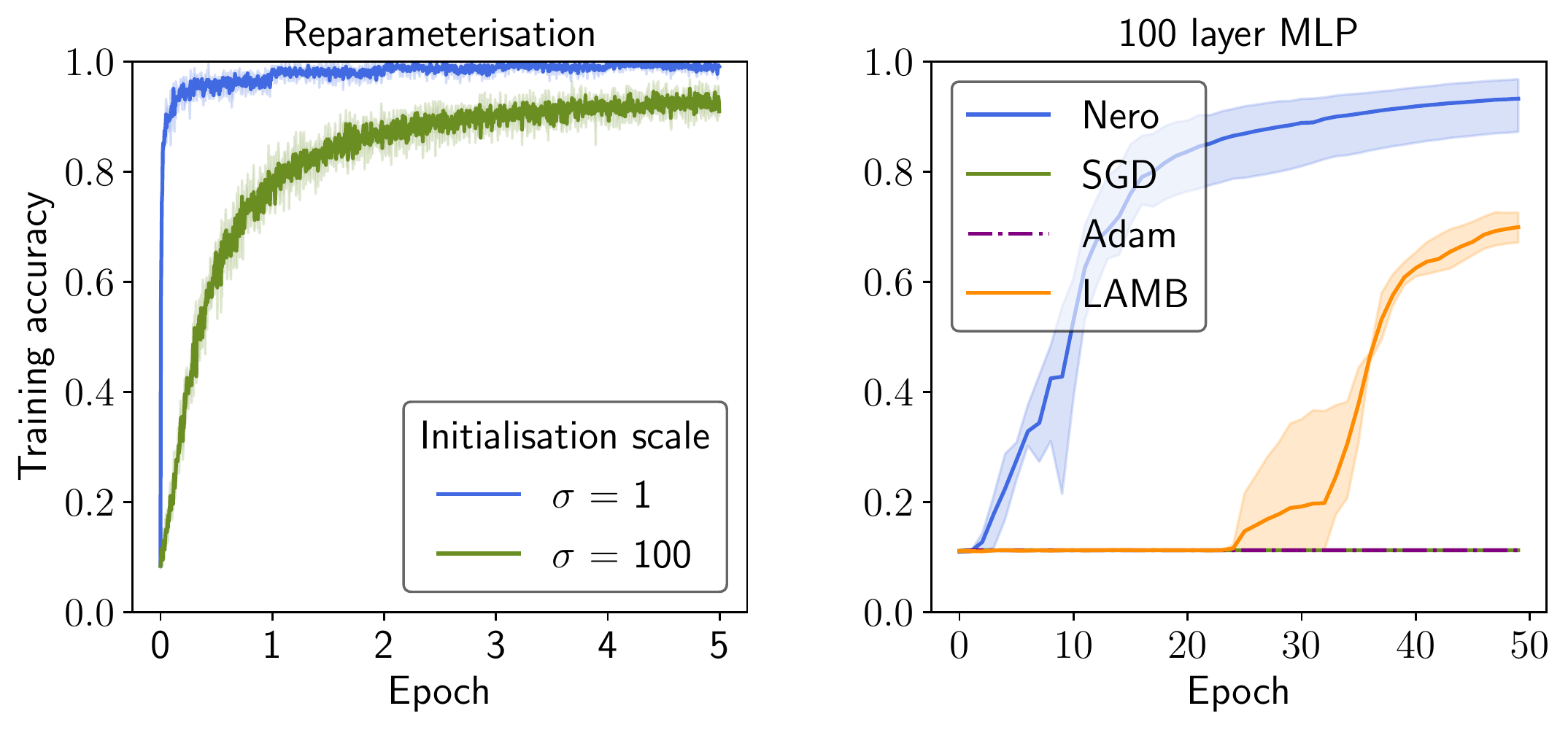}
    \vspace{-.75em}
    \caption{Left: Training a 5 layer perceptron normalised via reparameterisation (Equation \ref{eq:cwn}) on MNIST. For a fixed Adam learning rate, training is sensitive to the scale $\sigma$ of the raw weights $\widetilde{w}$. This motivates the different approach taken by Nero. Right: Using Nero to train a 100 layer perceptron---without batch norm or skip connections---to classify MNIST.}
    \label{fig:mnist}
\end{figure}

%% file: figures/table.tex
\begin{table*}[t]
    \centering
    \resizebox{.98\textwidth}{!}{%
    \begin{tabular}{cccccccccccc}
    \toprule
     \textbf{Task} & \textbf{Dataset} & \textbf{Model} & \textbf{Metric} $\bm{(\updownarrows)}$ & \textbf{Nero} & \textbf{SGD} & \textbf{Adam} & \textbf{LAMB} & \textbf{Nero} $\bm{\eta}$ & \textbf{SGD} $\bm{\eta}$ & \textbf{Adam} $\bm{\eta}$ & \textbf{LAMB} $\bm{\eta}$\\ 
     \midrule
     
     cGAN & CIFAR-10 & BigGAN-like & FID ($\downarrow$)& 
     $\bm{15.43 \pm 0.37}$  & $33.06 \pm 0.42$  & $23.42 \pm 0.85$ & \underline{$16.32 \pm 0.23$} & 0.01 & 0.01 & 0.0001 & 0.01\\
     
     Classification & CIFAR-10 & VGG11 & Top-1 Error ($\downarrow$) & 
     $\bm{11.16 \% \pm 0.17}$  & \underline{$12.61 \% \pm 0.21$}  & $12.86 \% \pm 0.34$ & $13.66 \% \pm 0.05$ &
     0.01 & 0.1 & 0.001 & 0.01\\

     Classification & CIFAR-10 & ResNet-18 & Top-1 Error ($\downarrow$) & 
     $\bm{5.75 \% \pm 0.07}$  & $7.75 \% \pm 0.17$  & \underline{$5.93 \% \pm 0.19$} & $6.46 \% \pm 0.12$ &
     0.01 & 0.1 & 0.01 & 0.1\\

     Language Model & Wikitext-2 & Transformer & Perplexity ($\downarrow$) & 
     $\bm{172.99 \pm 0.51}$ &  $181.76 \pm 0.49$  & \underline{$178.05 \pm 0.96$}  & $200.54 \pm 0.53$ &
     0.01 & 1.0 & 0.0001 & 0.01\\
     
     Translation & WMT16 En--De & Transformer & Perplexity ($\downarrow$) &
     $\bm{11.35 \pm 1.20}$ &  $92.40 \pm 89.48$  & \underline{$12.63 \pm 0.34$}  & $16.36 \pm 0.29$ &
     0.001 & 0.0001 & 0.0001 & 0.01\\
     
     PPO & Atari Pong & vanilla CNN & Reward ($\uparrow$) & 
     $\bm{20.62 \pm 0.05}$ &  $11.99 \pm 8.65$  & \underline{$15.92 \pm 3.40$}  & $-19.46 \pm 0.10$ & 
     0.01 & 0.1 & 0.0001 & 0.001\\
     \bottomrule
    \end{tabular}%
    }
    \captionsetup{justification=centering}
    \caption{Validation results for the best learning rate $\eta$. The best result is shown in bold, while the runner-up is underlined.}%
    \label{table:results}
\end{table*}

%% file: figures/cGAN.tex
\begin{figure}[H]
    \centering
    \includegraphics[height=0.45\linewidth]{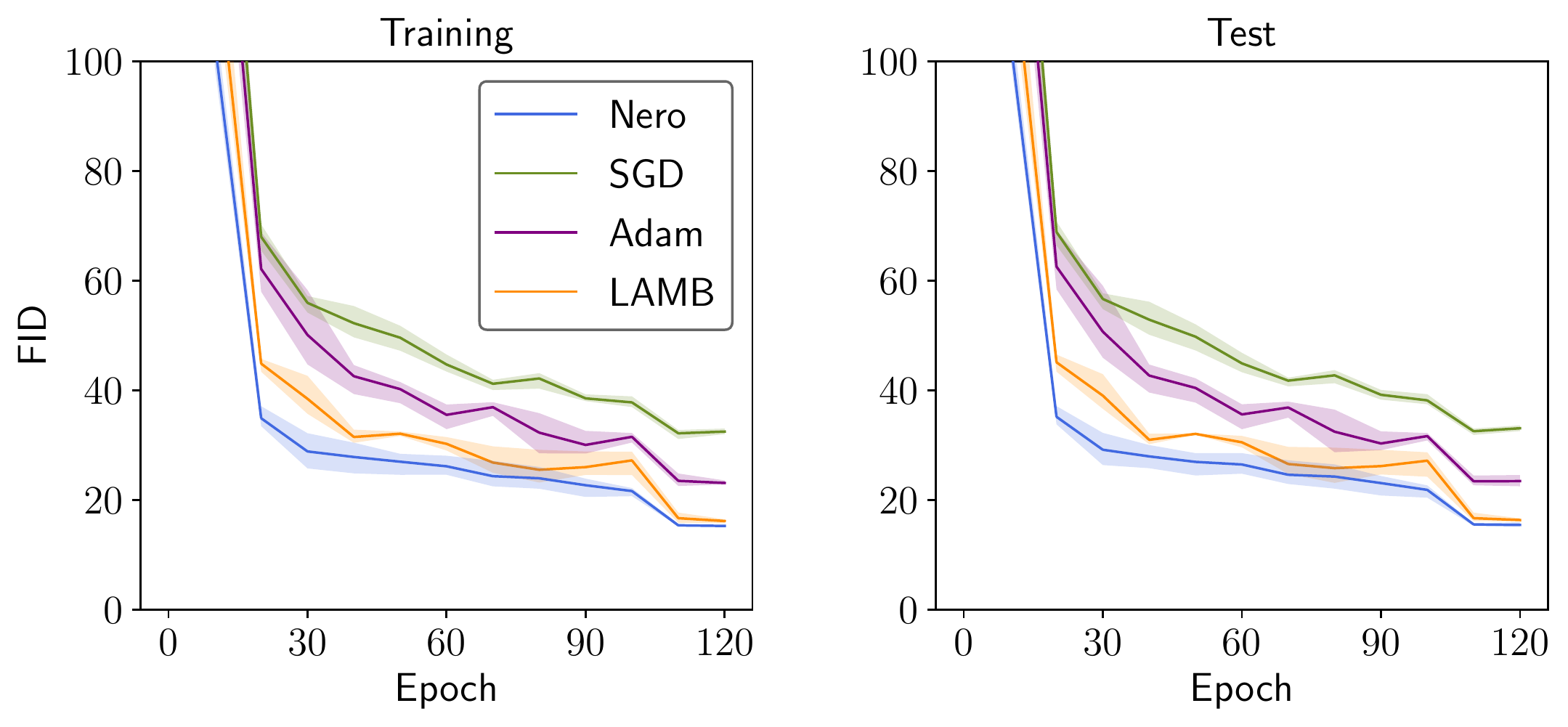}
    \vspace{-.75em}
    \caption{Class-conditional GAN training on CIFAR-10. Equal learning rates were used in the generator and discriminator. The Fréchet Inception Distance \citep[FID]{ttur} measures the distance between the sample statistics of real and fake data as represented at a deep layer of a pre-trained image classifier.
    }
    \label{fig:cgan}
\end{figure}

%% file: figures/cifar.tex
\begin{figure}[H]
    \centering
    \includegraphics[height=0.45\linewidth]{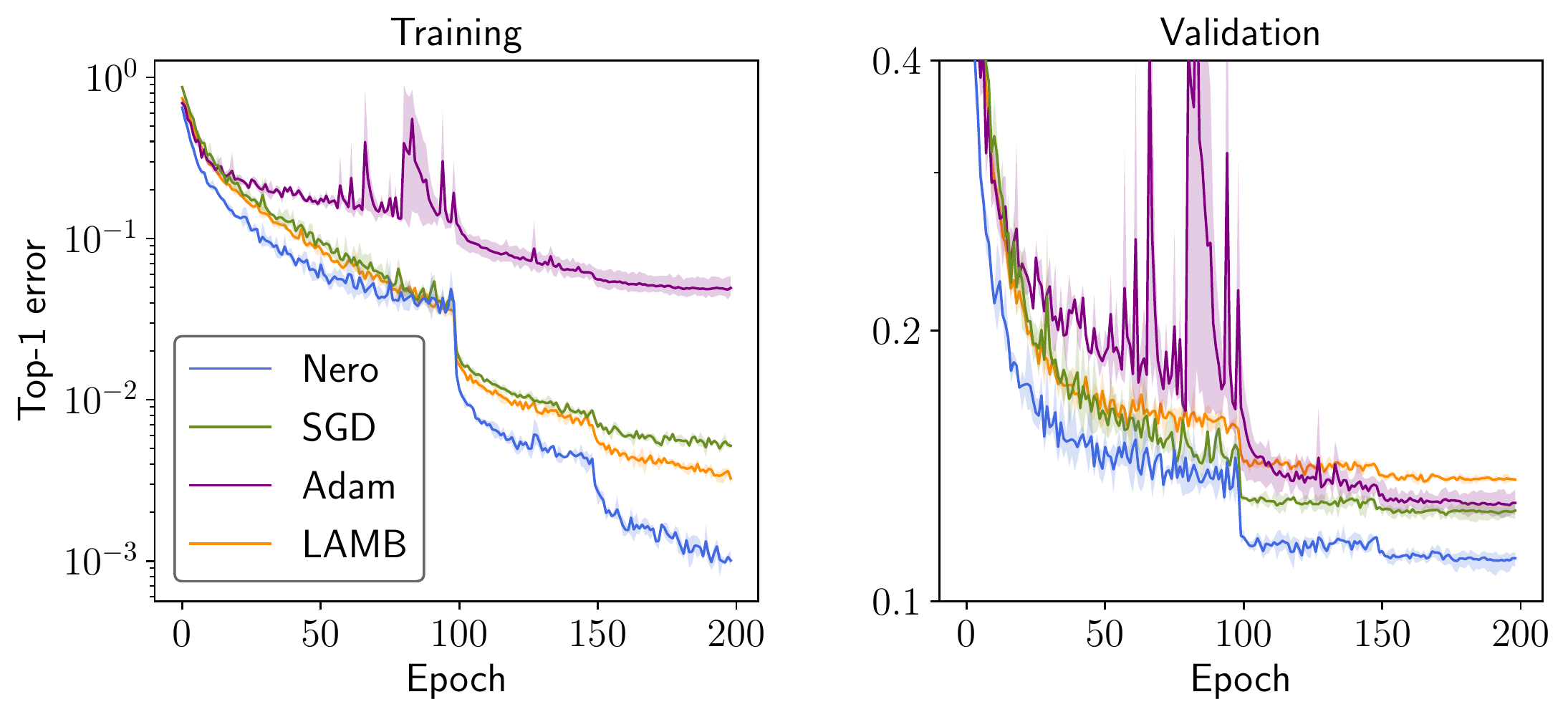} 
    \includegraphics[height=0.45\linewidth]{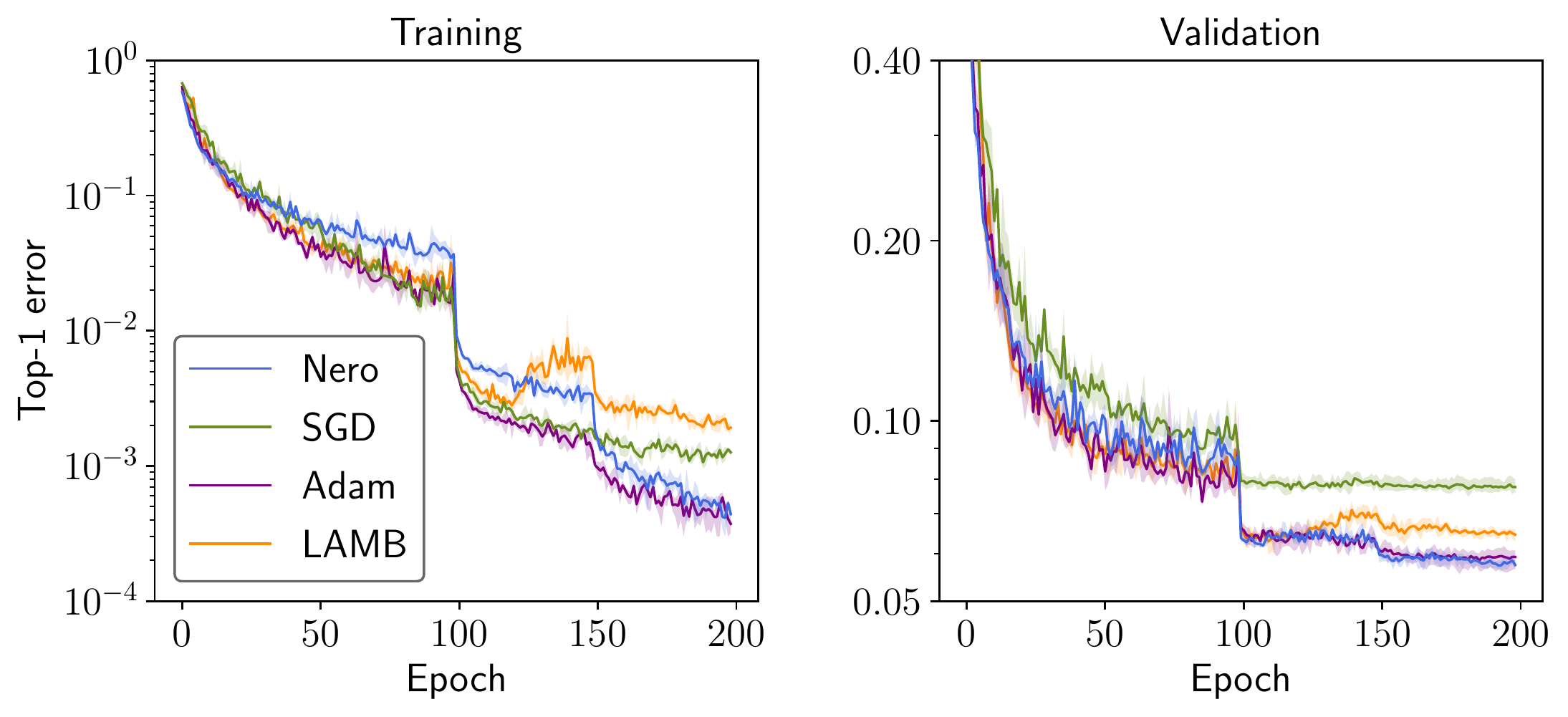}
    \vspace{-.75em}
    \caption{CIFAR-10 classification. Top: performance of a vanilla, convolutional VGG-11 network. Bottom: performance of a batch-normalised, residual ResNet-18 network.}
    \label{fig:cifar}
\end{figure}

%% file: figures/wikitext.tex
\begin{figure}[H]
    \centering
    \includegraphics[height=0.45\linewidth]{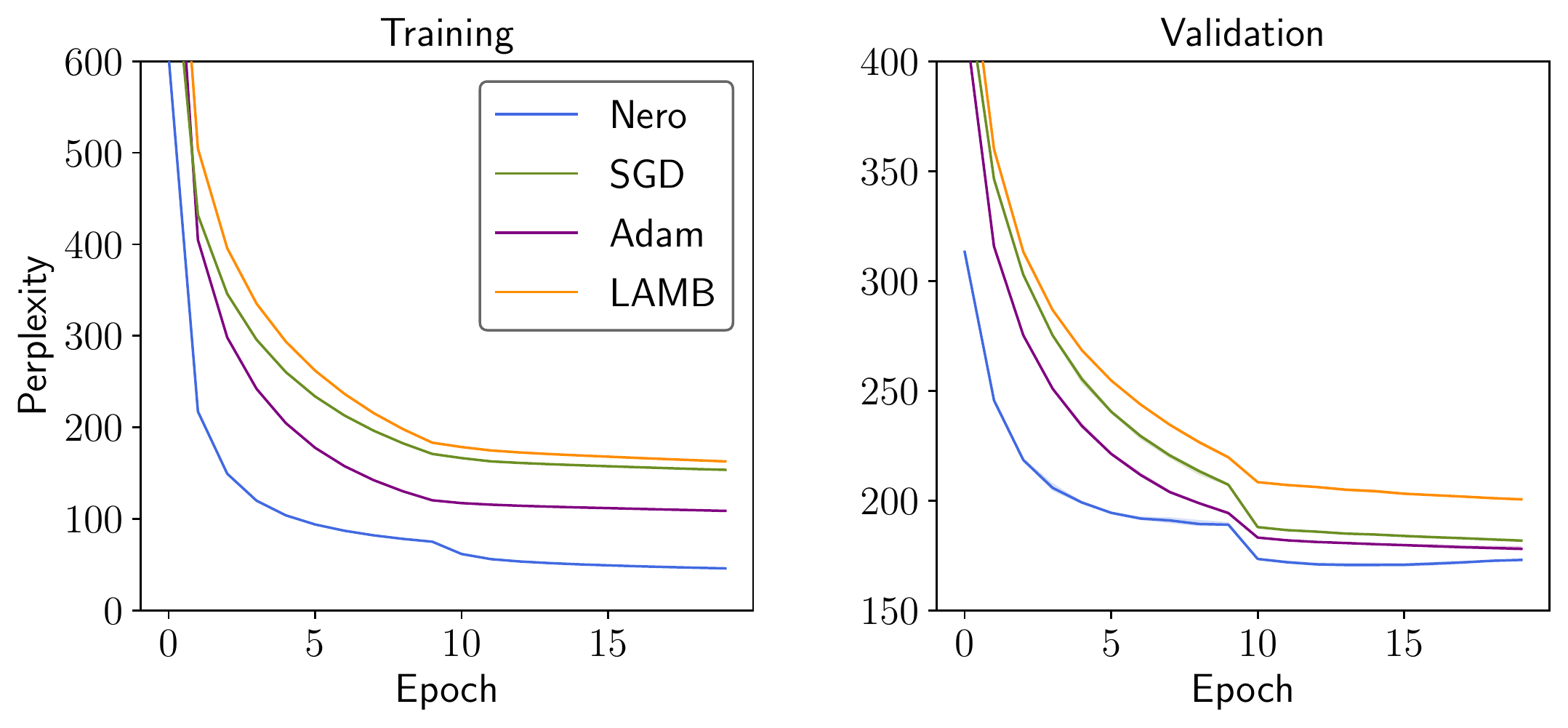}
    \vspace{-.75em}
    \caption{Training a language model on the Wikitext-2 dataset. A small transformer network was used, composed of 19 tensors. Nero achieved the best anytime performance.}
    \label{fig:wikitext}
\end{figure}

%% file: figures/translation.tex
\begin{figure}[H]
    \centering
    \includegraphics[height=0.45\linewidth]{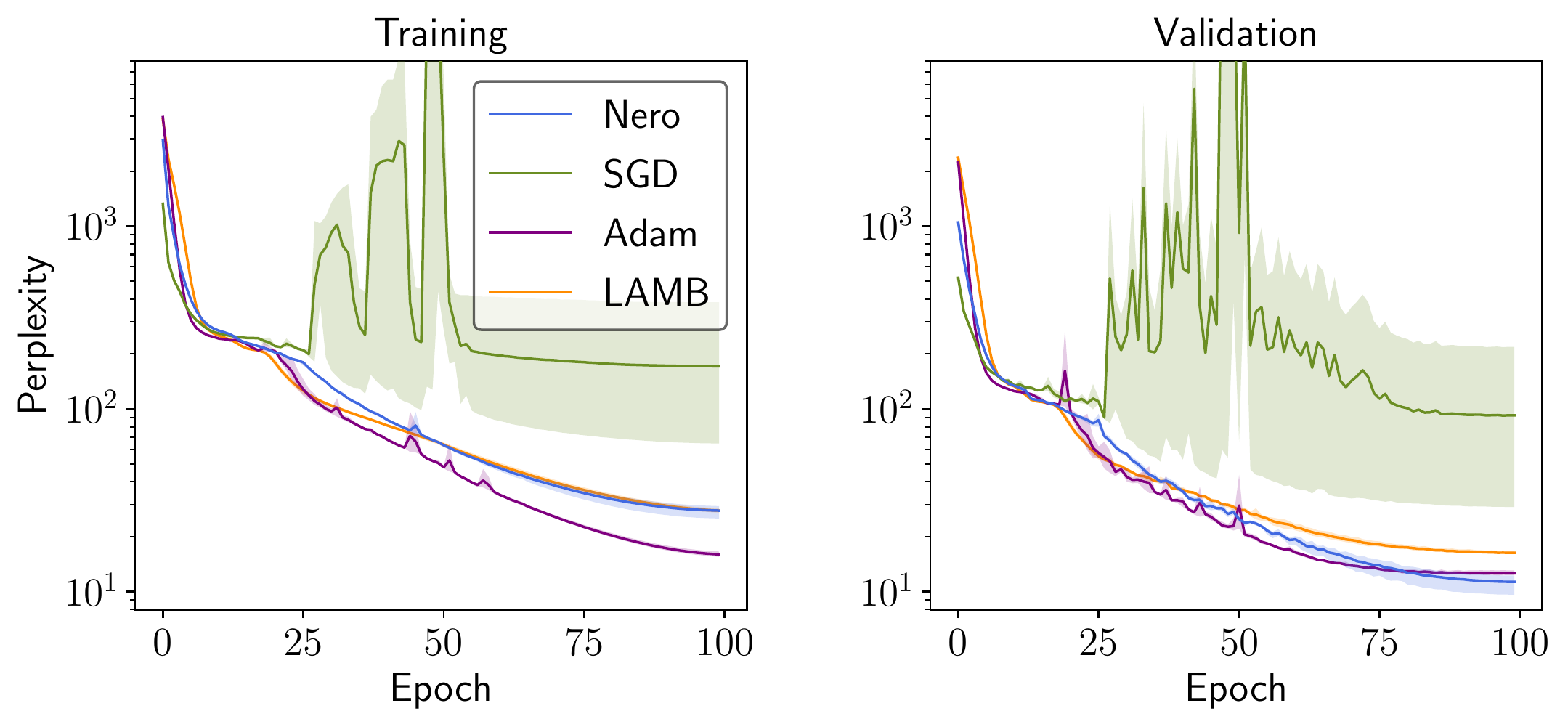}
    \vspace{-.75em}
    \caption{Training an English--German translation model on WMT16. A larger transformer network was used, composed of 121 tensors. The optimisers with gradient normalisation---Nero, Adam, and LAMB---performed best in training this model. Training with SGD was unstable and led to significantly worse perplexity. }
    \label{fig:translation}
\end{figure}

%% file: figures/ppo.tex
\begin{figure}[H]
    \centering
    \begin{tabular}{ c }
        \includegraphics[height=0.45\linewidth]{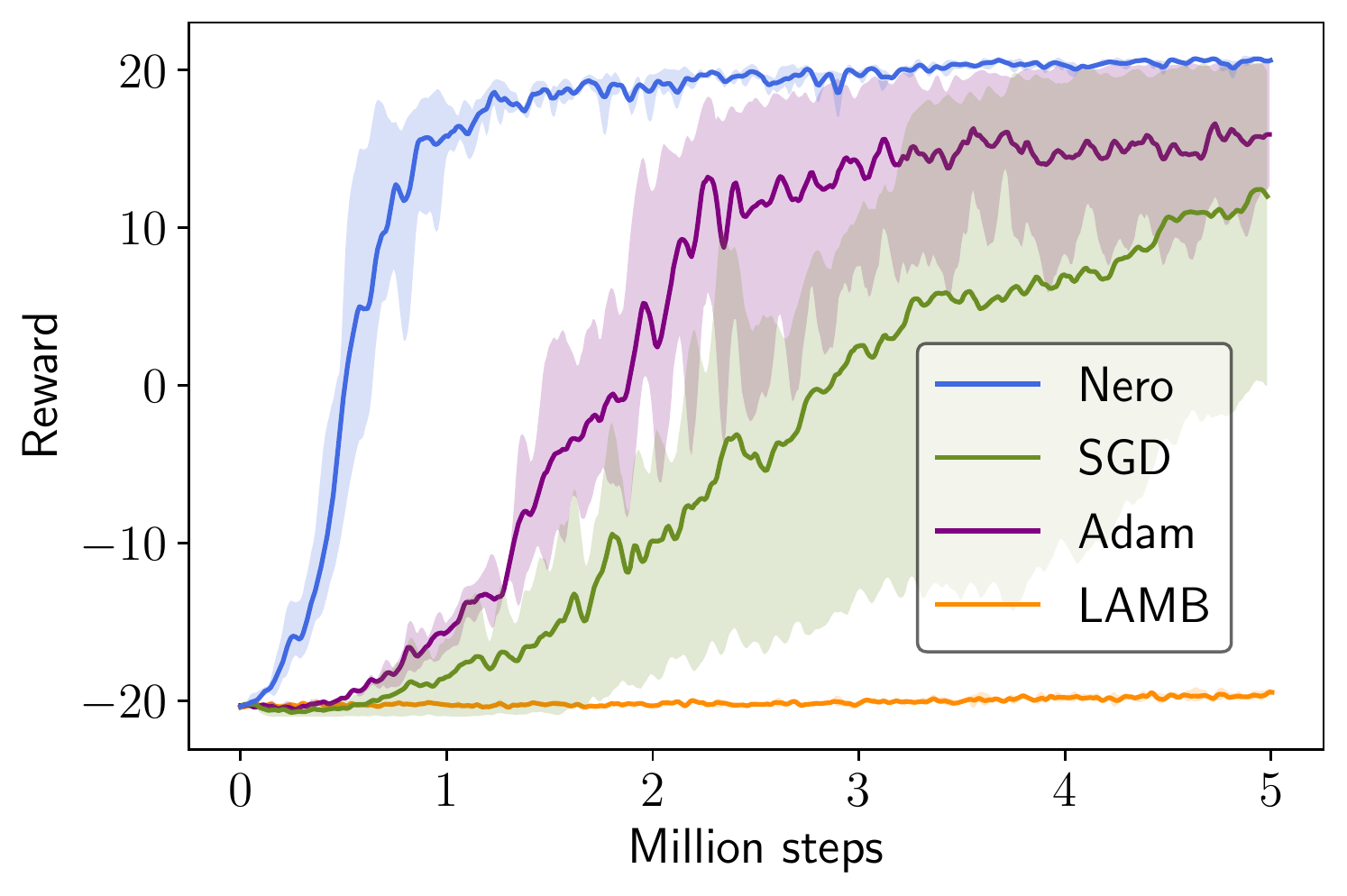} \\
    \end{tabular}
    \vspace{-.75em}
    \caption{Training a policy network to play Pong. Proximal Policy Optimisation (PPO) was used. Pong's reward is bounded between $\pm 21$. While investigating LAMB's failure to train the policy network, it was discovered that adjusting the $\beta_1$ momentum hyperparameter from 0 to 0.9 improved LAMB's performance.}%
    \label{fig:ppo}
\end{figure}

%% file: figures/imagenet.tex
\begin{figure}[H]
    \centering
    \includegraphics[height=0.45\linewidth]{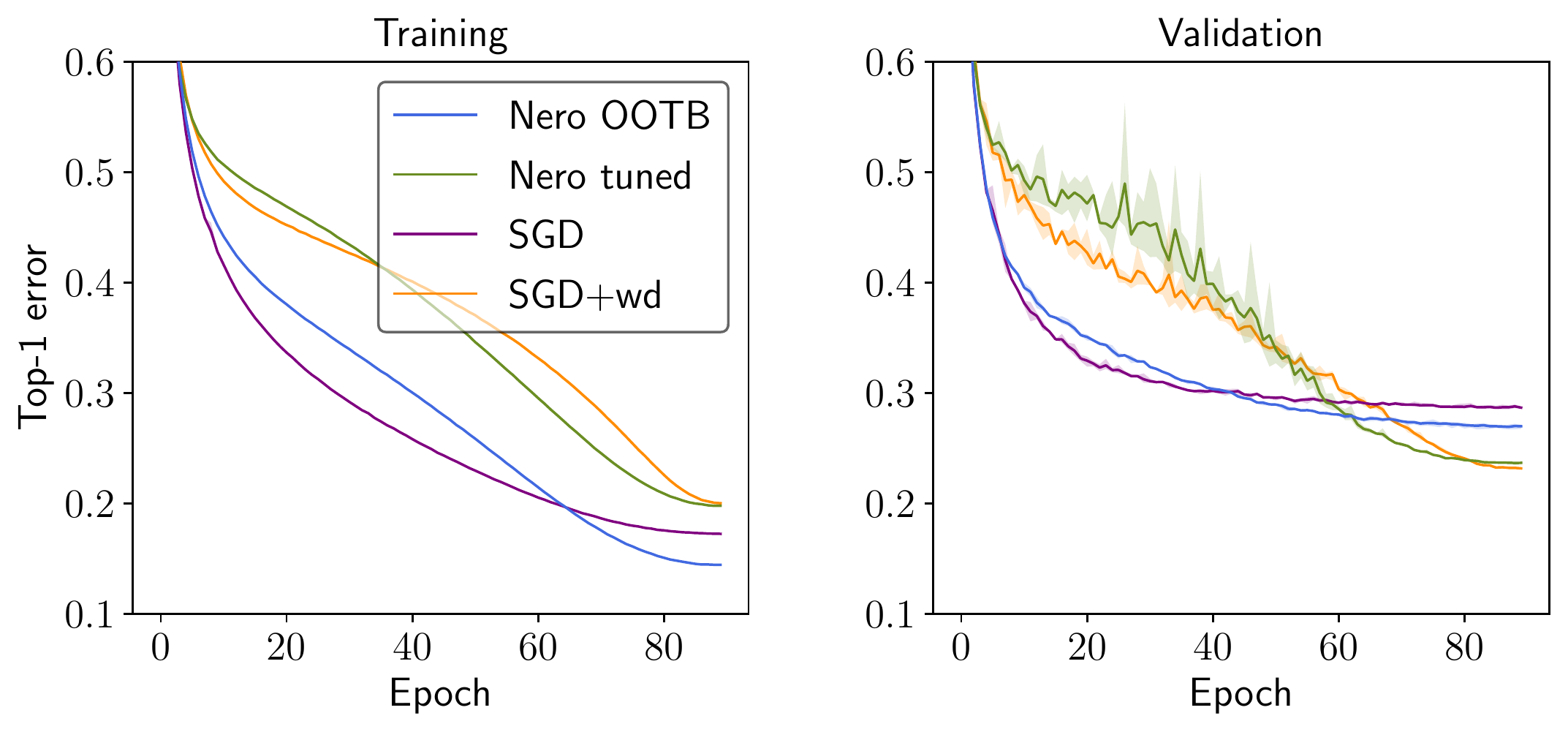} 
    \vspace{-.75em}
    \caption{Training a ResNet-50 network to classify the ImageNet dataset. \textit{Nero OOTB} (out-of-the-box) achieved the best training performance but overfit compared to SGD with weight decay.
    \textit{Nero tuned}---which most importantly regularised batch norm gains towards one---recovered most of the lost performance.
    }
    \label{fig:imagenet}
\end{figure}

%% file: section/8-discuss.tex
\section{Discussion and Future Work}

While the focus of this paper has been on motivating Nero and demonstrating its practical advantages, this section will discuss some of the more theoretical issues concerning convergence and generalisability of the trained network, as well as possible directions for future work.

\subsection{Convergence}

Convergence analyses of first order optimisation algorithms typically rely on a model of \textit{smoothness} of the loss function. One of the most commonly encountered models is \textit{Lipschitz smoothness} \citep{Nesterov2004IntroductoryLO}, although \citet{Sun2019OptimizationFD} points out that this model is somewhat ill-suited to neural networks.

Nero was motivated in Section \ref{sec:theory} based on a notion of \textit{layerwise relative smoothness} called \textit{deep relative trust} (Definition \ref{def:drt}). Deep relative trust attempts to directly model the smoothness of neural network loss functions, and may be used to derive formal convergence guarantees---see for instance \citep[Lemma 2]{fromage}.

Yet the empirical success of Nero---which rotates \textit{neurons} through fixed angles---might suggest that there is something missing in a notion of \textit{layerwise} relative smoothness. Indeed the success of Nero might suggest that neural networks are better characterised by a notion of \textit{per-neuron angular smoothness}. For instance, one might surmise that solutions returned by Nero satisfy a notion of angular robustness, as expressed in the following definition:
\begin{definition}\label{def:ang}
    A solution that attains zero training error is \textit{$\alpha$-robust} if all neurons may be simultaneously and arbitrarily rotated by up to angle $\alpha$ without inducing an error.
\end{definition}
In other words, Definition \ref{def:ang} suggests measuring the sharpness/flatness of converged solutions in terms of an angular parameter $\alpha$. Such a notion plays a role in generalisation theory, as will be seen in the next section.

\subsection{Generalisation}

The results in this paper may have a bearing on the generalisation theory of neural systems---an area of research that is still not settled. Consider the following hypothesis:
\begin{hypothesis}\label{h:norm}
    Deep learning generalises because SGD is biased towards solutions with small norm.
\end{hypothesis}
\vspace{-.5em}
This hypothesis is well-known, and is alluded to or mentioned explicitly in many papers \citep{NIPS2017_81b3833e,Zhang2017UnderstandingDL,Bansal2018MinnormTA,ADVANI2020428}.

But in light of the results in Table \ref{table:results}, Hypothesis \ref{h:norm} encounters some basic problems. First, for some tasks---such as the GAN and translation experiment---SGD simply performs very poorly. And second, Nero is able to find generalising solutions even when the norm of the network is constrained. For instance, the VGG-11 network and the Wikitext-2 transformer model have no gain parameters so, under Nero, the norm of the weights (though not the biases) is fixed and cannot be ``adapting to the data complexity''.

Then it seems right to consider an alternative theory:
\begin{hypothesis}\label{h:bayes}
    \textit{Deep learning generalises because the space of networks that fit the training data has large measure.}
\end{hypothesis}
\vspace{-.5em}
This hypothesis is essentially the PAC-Bayesian generalisation theory \citep{McAllester, Langford01boundsfor} applied to deep learning. \citet{valle-perez2018deep} have developed this line of work, proving the following result:

\begin{restatable}[Realisable PAC-Bayes]{theorem}{pacbayes}
\label{thm:pacbayes}
First, fix a probability measure $\Probe$ over the weight space $\Omega$ of a classifier. Let $S$ denote a training set of $n$ iid datapoints and let $V_S\subset\Omega$ denote the version space---that is, the subset of classifiers that fit the training data. Consider the population error rate $0\leq \eps(w)\leq 1$ of weight setting $w\in\Omega$, and its average over the version space $\eps(V_S) := \Expect_{w\sim\Probe}[\eps(w)|w\in V_S]$. Then, for a proportion $1-\delta$ of random draws of the training set $S$,
\begin{equation}\label{eq:pac}
     \eps(V_S) \leq \ln \frac{1}{1-\eps(V_S)}\leq \frac{\ln\frac{1}{\Probe[V_S]} + \ln \frac{2n}{\delta}}{n-1}.
\end{equation}
\end{restatable}
The intuition is that for a larger measure of solutions $\Probe[V_S]$, less information needs to be extracted from the training data to find just \textit{one} solution, thus memorisation is less likely. 

A simple bound on $\Probe[V_S]$ is possible based on this paper's connection between optimisation and architecture, since the problem is reduced to hyperspherical geometry.
Consider a balanced network (Definition \ref{def:bal-net}) composed of $m$ neurons each with fan-in $d$. Then the optimisation domain is isomorphic to the Cartesian product of $m$ hyperspheres:
\begin{equation*}
    \Omega \cong \underbrace{\Sph^{d-2}\times ... \times \Sph^{d-2}}_{m \text{ times}},
\end{equation*}
while $\Probe$ can be fixed to the uniform distribution on $\Omega$.

Next, suppose that the version space consists of $K$ non-intersecting $\alpha$-robust solutions (Definition \ref{def:ang}). Geometrically, an $\alpha$-robust solution is the product of $m$ hyperspherical caps. Thus the measure of the version space satisfies:
\begin{equation}\label{eq:vol}
    \textstyle
    \Probe[V_S] = K \cdot \Probe[\mathrm{cap}_{d-2}(\alpha)]^m \geq \frac{K}{2^m}\sin^{m(d-2)}\frac{\alpha}{2},
\end{equation}
where $\mathrm{cap}_{d-2}(\alpha)$ denotes an $\alpha$-cap of $\Sph^{d-2}$, and the inequality follows from \citep[Lemma 2.3]{Ball1997AnEI}. Combining Inequality \ref{eq:vol} with Inequality \ref{eq:pac} yields the following generalisation bound for neural networks:
\begin{align*}
    \eps(V_S) &\leq \frac{m \ln 2 + m(d-2) \ln \frac{1}{\sin \frac{\alpha}{2}} + \ln \frac{2n}{\delta}- \ln K}{n-1}.
\end{align*}
Focusing on the dominant terms, the bound suggests that the average test error $\eps(V_S)$ over the space of solutions $V_S$ is low when the number of datapoints $n$ exceeds the number of parameters $md$ less the entropy $\ln K$ of the multitude of distinct solutions. The theory has two main implications:
\begin{enumerate}%
    \item In the ``over-parameterised'' regime $md \gg n$, generalisation can still occur if the number of distinct solutions $K$ is exponential in the number of parameters $md$. In practice, $\ln K$ might be increased relative to $md$ by constraining the architecture based on the symmetries of the data---e.g. using convolutions for image data.
    \item All else equal, solutions with larger $\alpha$-robustness may generalise better. In practice, $\alpha$ might be increased by regularising the training procedure \citep{foret2021sharpnessaware}.
\end{enumerate}
Future work might investigate these ideas more thoroughly.

\subsection{Finer-Grained Architectural Awareness}

The experiments in this paper suggest that Nero performs well across a wide variety of neural architectures with heterogeneous building blocks---including convolution, attention, embeddings and normalisation layers with gains and biases. Yet the theoretical insights were derived primarily by considering simple neuronal building blocks. A more fine-grained study of different neural network components might yield both improved theoretical understanding and better empirical performance. Indeed this might lead to a more faithful realisation of \textit{neural architecture aware optimisation}.

\newpage

On a related note, it seems there is scope for new techniques that perform \textit{neural architecture aware regularisation}. The results in Section \ref{sec:imagenet} demonstrated the empirical effectiveness of one such technique: regularising gain parameters towards one. Regularising gains towards one is arguably more interpretable than applying weight decay to a flattened vector of all the network weights, and the authors of this paper found it significantly easier to tune.

More generally, in any setting where a deep learning technique operates on a flattened vector of all the network weights, it seems there is a good chance that the technique may be improved by accounting for the neural architecture.

\section{Conclusion}

This paper has proposed the Nero optimiser based on a combined study of optimisation and neural architecture. Nero pairs two ingredients: (1) projected gradient descent over the space of balanced networks; and (2) per-neuron relative updates. Taken together, a Nero update \textit{turns} each neuron through an angle set by the learning rate.

Nero was found to have strong \textit{out-of-the-box} performance. In almost all the experiments in this paper---spanning GAN training, image classification, natural language processing and reinforcement learning---Nero trained well using its default hyperparameter settings. The two exceptions were the 100 layer MLP and the WMT16 En--De transformer, for which Nero required a reduced learning rate of $\eta=0.001$. Thus Nero has the potential to accelerate deep learning research and development, since the need for time and energy intensive hyperparameter search may be reduced.

%% file: section/99-app.tex
\section{Experimental Details}\label{app:expt}

All code is available at \href{https://github.com/jxbz/nero}{\texttt{github.com/jxbz/nero}}. This appendix records important details of the implementations.

\paragraph{MNIST classification} These experiments used a multilayer perceptron (MLP) network. An $L$-layer architecture consisted of $(L-1)$ layers of dimension $784\times784$ followed by an output layer of dimension $784\times10$. A ``scaled relu'' nonlinearity was used, defined by $\phi(x) := \sqrt{2}\cdot\max(0,x)$. The factor of $\sqrt{2}$ was motivated by \textit{Kaiming init} \citep{he2015delving} and was not tuned. The reparameterisation experiment used $L=5$ layers and trained for 5 epochs without learning rate decay. The very deep MLP used $L=100$ layers and trained for 50 epochs with the learning rate decayed by a factor of 0.9 at the end of every epoch, and with the initial learning rate tuned over $\{0.0001, 0.001, 0.01, 0.1\}$. Training took place on an unknown Google Colab GPU. On an NVIDIA Tesla P100 GPU, the 5-layer MLP took $\sim 1$ minute to train and the 100-layer MLP took $\sim 30$ minutes.

\paragraph{CIFAR-10 cGAN} Equal learning rates were used in the generator and discriminator. The initial learning rate was tuned over \{0.0001, 0.001, 0.01, 0.1, 1.0\} for all optimisers. The networks were trained for 120 epochs, with the learning rate decayed by a factor of 10 at epoch 100. The momentum parameter in SGD and $\beta_1$ in Adam and LAMB were tuned over \{0.0, 0.9\}. Nero's $\beta$ and $\beta_2$ in Adam and LAMB were set to 0.999 without tuning. Training took around 3 hours on an NVIDIA RTX 2080Ti GPU.

\paragraph{CIFAR-10 classification} All models were trained for 200 epochs, with 5 epochs of linear learning rate warm-up and learning rate decay by a factor of 0.2 at epochs 100, 150 and 180. The initial learning rates were tuned over \{0.0001, 0.001, 0.01, 0.1, 1.0\}. Training was performed on an NVIDIA RTX 2080Ti GPU. Training time for the VGG-11 network was $\sim 1$ hour, and for ResNet-18 was $\sim 2$ hours.

Since the experiments in Figures \ref{fig:ablation} and \ref{fig:nml} were intended to probe the fundamental properties of optimisers rather than their performance under a limited tuning budget, a more fine-grained learning rate search was conducted. Specifically, the learning rates were tuned over \{0.01, 0.02, 0.04, 0.06, 0.08, 0.1\}. The best results are listed in the following table:

\input{figures/nml_table}

\paragraph{ImageNet classification} For training with SGD + momentum + weight decay, the initial learning was set to 0.1, momentum was set to 0.9 and weight decay was set to 0.0001. These settings follow \citet{he2016deep}. One epoch of linear learning rate warm-up was used, followed by 89 epochs of cosine annealing. The batch size was set to 400 for ResNet-50 to fit the GPU vRAM budget, and this was in the range known to yield good performance \citep{goyal2017accurate}. This paper's SGD implementation surpassed the target ImageNet top-1 accuracy of $76.3\%$ for ResNet-50 \citep{goyal2017accurate,lamb}. The training was distributed over four NVIDIA RTX 2080Ti GPUs, taking $\sim 45$ hours per training run. The total number of GPU hours for all ImageNet experiments in this paper was $\sim 1500$.

\paragraph{Wikitext-2 language model} The small transformer model was trained for 20 epochs, with the learning rate decayed by a factor of 0.1 at epoch 10. The initial learning rate was tuned over \{0.0001, 0.001, 0.01, 0.1, 1.0\}. The batch size was set to 20. Training on an NVIDIA RTX 2080Ti GPU took $\sim 15$ minutes.

\paragraph{WMT16 En--De translation} The large transformer model was trained for 100 epochs, with a linear warm-up from epoch 0 to 50, and linear annealing from epoch 50 to 100. The maximum learning rate was tuned over \{0.0001, 0.001, 0.01, 0.1, 1.0\}. A batch size of 128 was used. Training took $\sim 1$ hour on an NVIDIA RTX 2080Ti GPU.

\paragraph{Reinforcement learning} Hyperparameter settings followed \citet{pytorchrl}, except for the initial learning rate and the total number of environment steps. The number of steps was fixed to 5 million, and the initial learning rate was tuned over \{0.0001, 0.001, 0.01, 0.1, 1.0\}. The policy network combined convolutional image feature extractors with dense output layers. Training was performed on an NVIDIA RTX 2080Ti GPU, and the training time was $\sim 1.5$ hours.

%% file: figures/nml_table.tex
\begin{flushleft}
\resizebox{\columnwidth}{!}{%
\begin{tabular}{ccccc}
\toprule
 \textbf{Optimiser} & \textbf{Fix Mean} & \textbf{Fix Norm} & \textbf{Top-1 Error} & \textbf{Best} $\bm{\eta}$ \\
 \midrule
 Nero  &            &            & $   {12.17 \% \pm 0.08}$  & 0.02 \\
 Nero  & \checkmark &            & $   {11.99 \% \pm 0.14}$  & 0.01 \\
 Nero  &            & \checkmark & $   {10.03 \% \pm 0.24}$  & 0.02 \\
 Nero  & \checkmark & \checkmark & $\bm{ 8.61 \% \pm 0.22}$  & 0.02 \\ 
 Madam &            &            & $   {12.77 \% \pm 0.20}$  & 0.02 \\
 Madam & \checkmark & \checkmark & $   {12.60 \% \pm 0.12}$  & 0.06 \\
 LAMB  &            &            & $   {12.73 \% \pm 0.10}$  & 0.02 \\
 LAMB  & \checkmark & \checkmark & $   { 8.88 \% \pm 0.08}$  & 0.06 \\
 \bottomrule
\end{tabular}
}
\end{flushleft}